\documentclass[letterpaper, 10 pt, conference]{ieeeconf}  

\IEEEoverridecommandlockouts                              

\usepackage[vlined, ruled, boxed, linesnumbered]{algorithm2e}
\SetKwProg{Function}{}{}{}

\usepackage{graphicx}
\usepackage{graphics}
\usepackage{times}
\usepackage{amsmath}
\usepackage{amssymb}
\usepackage{float}
\usepackage{url}
\usepackage{subfigure}
\usepackage{caption}
\usepackage{multirow}
\usepackage{gensymb}
\usepackage{epstopdf}
\usepackage{wrapfig}
\usepackage{setspace} 
\usepackage[font={small}]{caption}

\title{\LARGE \bf
A Lexicographic Search Method for Multi-Objective Motion Planning
}

\author{Tixiao Shan and Brendan Englot
\thanks{T. Shan and B. Englot are with the Department of Mechanical Engineering, Stevens Institute of Technology, Castle Point on Hudson, Hoboken NJ 07030 USA, {\tt\small \{TShan3, BEnglot\}@stevens.edu}. }%
}

\begin{document}

\maketitle
\thispagestyle{empty}
\pagestyle{empty}

\begin{abstract}

We propose a novel method for multi-objective motion planning problems by leveraging the paradigm of \textit{lexicographic optimization} and applying it for the first time to graph search over probabilistic roadmaps. The competing resources of interest are penalized hierarchically during the search. Higher-ranked resources cause a robot to incur non-negative costs over the paths traveled, which are occasionally zero-valued. This is intended to capture problems in which a robot must manage resources such as visibility of threats, availability of communications, and access to valuable measurements. This leaves freedom for tie-breaking with respect to lower-priority resources; at the bottom of the hierarchy is a strictly positive quantity consumed by the robot, such as distance traveled, energy expended or time elapsed. We compare our method with two other multi-objective approaches, a naive weighted sum method and an expanded graph search method, demonstrating that a lexicographic search can solve such planning problems efficiently without a need for parameter-tuning in unintuitive units. The proposed method is also demonstrated on hardware using a laser-equipped ground robot.



\end{abstract}

\section{INTRODUCTION}

Multi-objective motion planning has been an area of interest in robotics for many years. Continuous multi-objective motion planning in two and three dimensions has been achieved by gradient descent, paired with sampling the Pareto front to identify feasible solutions under added constraints \cite{mitchell2003}. Genetic algorithms \cite{castillo2007}, \cite{vadakkepat2000} and dynamic programming \cite{hansen1980}-
\cite{martins1984} have also been applied to solve multi-objective motion planning problems. Early work on multi-objective planning over configuration space roadmaps \cite{lavalle1998} has been accomplished by methods that recover Pareto fronts from probabilistic roadmaps (PRMs) \cite{clawson2015}, \cite{ding2014}. In pursuit of solutions that can be produced quickly, preferably in real-time, and applied to problems of high dimension, sampling-based motion planning algorithms such as the PRM \cite{kavraki1996}, the rapidly-exploring random tree (RRT) \cite{lavalle2001}, and their optimal variants PRM*, RRT*, and rapidly-exploring random graphs (RRG) \cite{karaman2011} have been adapted to solve a variety of multi-objective motion planning problems. Such approaches have typically considered the tradeoff between a resource such as time, energy, or distance traveled and a robot's information gathered \cite{hollinger2014}, localization uncertainty \cite{bopardikar2015}, \cite{luders2014}, collision probability \cite{roy2011}, clearance from obstacles \cite{kim2003}, adherence to rules \cite{reyes2013},  and exposure to threats \cite{clawson2015} and other generalized representations of risk \cite{devaurs2014,jaillet2010}. 

We consider problems in which two or more competing resources are penalized \textit{hierarchically}. The higher-priority resources assume non-negative costs over robot paths, and are frequently zero-valued. This is intended to capture problems in which robots must manage resources such as visibility to threats, collision risk, availability of communications or access to valuable measurements, which are present in some regions of the environment, and absent in others. For example, although some algorithms seek to manage collision risk by adhering closely to the medial axis of the free space \cite{denny2014}, \cite{yeh2014}, this is often overly conservative. Penalizing such risks only in the regions of the environment where collision is possible leaves freedom for tie-breaking with respect to a secondary resource, which may be a strictly positive quantity consumed by the robot, such as distance traveled, energy expended or time elapsed \cite{shan2015}. Such a problem fits nicely into the framework of \textit{lexicographic optimization}. 

The lexicographic method \cite{stadler1988} is the technique of solving a multi-objective optimization problem by arranging each of several cost functions in a hierarchy reflecting their relative importance. The objectives are minimized in sequence, and the evolving solution may improve with respect to every subsequent cost function if it does not worsen in value with respect to any of the former cost functions. Use of this methodology has been prevalent in the civil engineering domain, in which numerous regulatory and economic criteria often compete with the other objectives of an engineering design problem. Variants of the lexicographic method have been used in land use planning \cite{veith2003}, for vehicle detection in transportation problems \cite{sun1999}, and in the solution of complex multi-objective problems, two criteria at a time \cite{engau2007}. 

Among the benefits of such an approach is the potential for the fast, immediate return of a feasible solution that offers globally optimal management of the primary resource, in addition to locally optimal management of secondary resources in areas where higher-ranked resources are zero-valued. Due to the fact that the spatial regions in which resources are penalized can often be intuitively derived from a robot's workspace, using facts such as whether there is a line-of-sight to a known threat, or whether a robot is within range of communication or sensing resources, such an approach offers an intuitive means for managing the relative importance of competing cost functions, in which the user needs only to select the order in which the resources are penalized. This stands in contrast to methods that require tuning of additive weights on the competing cost functions \cite{reyes2013}, and robot motion planning methods that manage the relative influence of competing cost criteria using constraints, \cite{hollinger2014}-
\cite{roy2011}, \cite{devaurs2014}. Avoiding any potential struggles to recover feasible solutions under such constraints, the lexicographic motion planning problem is \textit{unconstrained} with respect to the resources of interest. 

The principal contribution of this paper is a search algorithm that can be utilized with a roadmap to solve multi-query, multi-objective motion planning problems quickly without parameter-tuning. The roadmap may be a deterministic construct such as a Voronoi or lattice graph, or it may be a PRM, PRM* or RRG. Online planning, re-planning or the navigation of multiple robots can be achieved by making repeated search calls to the roadmap. Upon ranking a robot's multiple objectives hierarchically, there is no need for further tuning of constraints. The proposed \textit{lexicographic search} is compared with (1) a method that uses a single compound cost function comprised of a weighted sum of the competing resources, and (2) a method that searches an \textit{expanded graph}, which uses increasing constraints for each graph. Section II defines the problem of interest, Section III describes and justifies the proposed solution, and Section IV discusses the complexity of the search algorithm. Section V presents a computational study comparing the proposed method with the competing expanded graph search and weighted sum methods. The algorithm's flexibility is demonstrated over bi-criteria and quad-criteria robot motion planning problems inspired by real-world autonomous navigation challenges, including bi-criteria results from robot hardware.

\section{PROBLEM DEFINITION}

Let $\mathcal{C}$ be a robot's configuration space. $x \in \mathcal{C}$ represents the robot\textquoteright s configuration. $\mathcal{C}_{obst} \subset \mathcal{C}$ denotes the set of configurations that are in collision with the obstacles. $\mathcal{C}_{free} = cl(\mathcal{C} \backslash \mathcal{C}_{obst})$, in which $cl()$ represents the closure of an open set, denotes the space that is free of collision in $\mathcal{C}$. We assume that given an initial configuration $x_{init} \in \mathcal{C}_{free}$, the robot must reach a goal state $x_{goal} \in \mathcal{C}_{free}$. Let a \textit{path} be a continuous function $\sigma : [ 0,1 ]  \rightarrow \mathcal{C}$ of finite length. Let $\Sigma$ be the set of all paths $\sigma$ in a given configuration space. A path $\sigma$ is collision-free and feasible if $\sigma \in \mathcal{C}_{free}$, $\sigma(0) = x_{init}$ and $\sigma(1) = x_{goal}$. The problem of finding a feasible path may be specified using the tuple $({C}_{free},x_{init},x_{goal})$. 

We assume the robot moves through $\mathcal{C}_{free}$ along paths obtained from a directed graph $G(V,E)$, with vertex set $V$ and edge set $E$. An edge $e_{ij} \in E$ is a path $\sigma_{i,j}$ for which $\sigma_{i,j}(0) = x_i \in V$ and $\sigma_{i,j}(1) = x_j \in V$. Two edges $e_{ij}$ and $e_{jk}$ are said to be linked if both $e_{ij}$ and $e_{jk}$ exist. A path $\sigma_{p,q} \in G$ is a collection of linked edges such that $\sigma_{p,q} = \{e_{p \; i_1},e_{i_1 i_2}, ..., e_{i_{n-1} i_n}, e_{i_n q}\}$.

We employ cost functions $f_k(\sigma)$, where $f_k: \Sigma \rightarrow \mathbb{R}_{0}^{+}$ maps a path $\sigma$ to a $k$th non-negative cost,  $k \in \{1,2,...,K\}$, and $K$ is the total number of costs in a multi-objective planning problem. These $K$ cost functions are applied to the problem of lexicographic optimization \cite{marler2004}, which may be formulated over collision-free paths as
\begin{align}
&\sigma^* = \underset{\sigma_{k}\in \mathcal{C}_{free}}{min}{f_{k}(\sigma) }\;\;\;\; \\
&subject \; to: \; f_{j}(\sigma)\leq f_{j}(\sigma_{j}^{*})\;\; \nonumber \\
&j=1,2,...,k-1, k>1; \; \; k=1,2,...,K. \nonumber
\end{align}
The formulation of the lexicographic method is adapted here (we refer the reader to the description from \cite{marler2004}, Section 3.3) to show cost functions that take collision-free paths as input. We also assume specifically that $f_K: \Sigma \rightarrow \mathbb{R}^{+}$, implying that ties never occur in the bottom level of the hierarchy. In one iteration of the procedure of Eq. (1), a new solution $\sigma^*$ will be returned if it does not increase in cost with respect to any of the prior cost functions $j < k$ previously examined. Necessary conditions for optimal solutions of Eq. (1) were first established by Rentmeesters \cite{rentmeesters1996}. Relaxed versions of this formulation have also been proposed, in which $f_{j}(\sigma_{k}) > f_{j}(\sigma_{j}^{*})$ is permitted, provided that $f_{j}(\sigma_{k})$ is no more than a small percentage greater in value than $f_{j}(\sigma_{j}^{*})$. This approach, termed the \textit{hierarchical method} \cite{osyczka1984}, has also been applied to multi-criteria problems in optimal control \cite{waltz1967}. 

\section{Algorithm Description}

We adapt Dijkstra's algorithm \cite{dijkstra1959} to perform a lexicographic graph search, which is detailed in Algorithm 1. Provided with a roadmap $G(V,E)$ and $x_{init}$ as inputs, a queue $X_{queue}$ is populated with the nodes of the roadmap (Line 1), and the algorithm initializes $K$ cost-to-come labels for each node  (Lines 2-4). Each of these labels describes the $k$th priority cost-to-come for a respective node, along the best path identified so far  per the ranking of cost functions in (1). In real-time applications of the search, $x_{init}$ is designated to be the closest configuration in the roadmap to the robot's current configuration.

\begin{algorithm}
\setstretch{0.8}
\caption{Lexicographic Search}
\KwIn{$G(V,E), x_{init}, x_{goal}$}
$X_{queue} \leftarrow \{V\}; \; X_{visited}\leftarrow \{\}$\;

\For{$k=1\:to\:K$}{
	$SetLabels_{k}(V,\infty)$\;
	$x_{init}.label_{k} \leftarrow 0$\;
}
$x_{init}.parent \leftarrow \{\}$\;

\While{$|X_{queue}| > 0$}{
	$X_{min}\leftarrow X_{queue}$\;	
   \For{$k=1\:to\:K$}{
		$X_{min}\leftarrow FindMinLabel_{k}(X_{min})$\;
		\If{$|X_{min}| = 1$}{
			$x_{i} \leftarrow X_{min}$\;
			$break$\;
		}
	}
	
	$X_{visited} \leftarrow X_{Visited} \cup x_{i}; \; X_{queue} \leftarrow X_{queue} \setminus x_{i}$\;
	
	\For{$\{x_{j} \; | \;  e_{ij}\in E\}$}{
		\For{$k=1\:to\:K$}{
			\uIf{$x_{j}.label_{k}>x_{i}.label_{k}+f_{k}(e_{ij})$}{
				$UpdateCost(x_{i},x_{j},e_{ij}, k)$\;
				$break$\;
			}
			\uElseIf{$x_{j}.label_{k}=x_{i}.label_{k}+f_{k}(e_{ij})$}{
				$continue$\;
			}
			\Else{	$break;$}
		}
	}
}

\end{algorithm}

\begin{algorithm}
\setstretch{0.8}
\caption{Update Cost Procedure}
\Function{$\textbf{UpdateCost}(x_{i},x_{j},e_{ij}, n)$}{
	\For{$k=n\:to\:K$}{
		$x_{j}.label_{k} \leftarrow x_{i}.label_{k}+f_{k}(e_{ij});$
	}
	$x_{j}.parent \leftarrow x_{i};$
}

\end{algorithm}

In each iteration of the algorithm's while loop, the $FindMinLabel_k()$ operation returns the set of configurations that share the minimum $k$th priority cost-to-come from among the nodes provided as input (Line 9). If $X_{min}$ contains more than one configuration, lower-priority costs for the nodes in  this set are examined until the set $X_{min}$ contains a single node, whose neighbors are examined in detail. The selected node is designated $x_i$ (Line 11). Node $x_{i}$ is then used, if possible, to reduce the costs-to-come associated with neighboring nodes $x_{j}$, if edge $e_{ij}$ exists. If the $k$th priority cost from $x_{init}$ to $x_{j}$ via $x_{i}$ is lower than the current cost, $x_{j}.label_{k}$, the costs-to-come of $x_{j}$ are updated by choosing $x_{i}$ as its new parent, per the function $UpdateCost()$ (Line 17). This function is detailed in Algorithm 2; when it is called, the labels of node $x_j$ are updated. If, however, the $k$th priority cost from $x_{init}$ to $x_{j}$ via $x_{i}$ is tied with the current cost, $x_{j}.label_{k}$ (Line 19), then Algorithm 1 proceeds to the lower-priority cost $k+1$ and evaluate the potential $(k+1)$th priority cost-to-come improvements at $x_j$ by traveling via $x_i$. To reduce the likelihood of end-stage ties, the lowest-priority cost $K$ is assumed to be strictly positive over all paths in the configuration space. 

\begin{figure}[ht]
\centering
\includegraphics[width=0.25\textwidth]{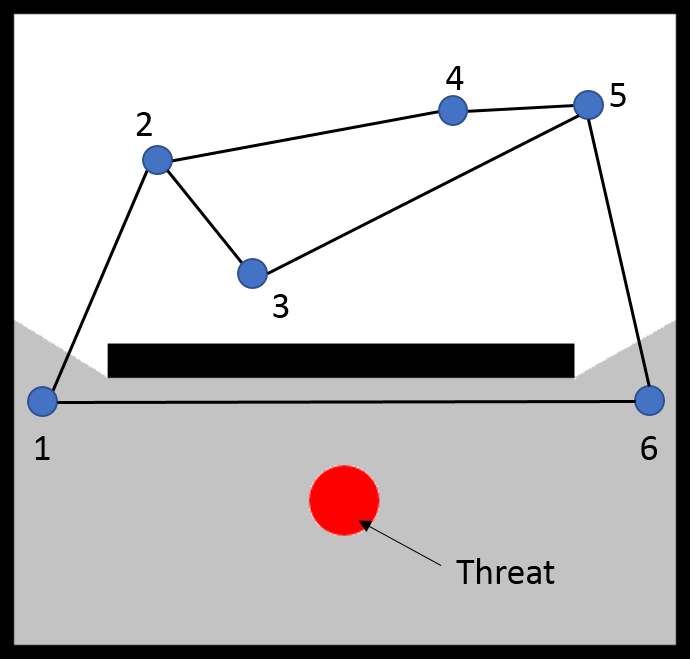}
\caption{A bi-criteria path planning problem well-suited to lexicographic search. A threat is colored red, an obstacle is colored black, and a penalty region, in which the threat has a clear line-of-sight to the robot, is colored gray. A risk penalty is incurred as a linear function of distance traveled by the robot in the threat's visibility polygon. Elsewhere, the penalty is zero and distance traveled is used to break ties.}
\vspace{-0mm}
\end{figure}

Just as the problem formulation in Equation (1) only allows improvements to a solution's secondary cost when it does not adversely impact a higher-priority cost, the proposed search method only allows improvements to be made in secondary costs when ties occur with respect to higher-priority costs. The single-source shortest paths solution produced by Algorithm 1 would take on the same primary cost whether or not these improvements are performed, but the occurrence of ties allows us to opportunistically address auxiliary cost functions in the style of lexicographic optimization. 

The example given in Figure 1 is designed to illustrate the mechanics of the proposed search method. Our goal is to find the path of minimum cumulative exposure to risk during a robot's travel from vertex 1 to vertex 6, and to minimize distance traveled as a secondary objective. Since portions of the edges $e_{12}$, $e_{16}$ and $e_{56}$ lie in the penalty region where a known threat has an unobstructed line-of-sight to the robot, the risk costs $f_1(e_{12})$, $f_1(e_{16})$ and $f_1(e_{56})$ are non-zero. On the other hand, edges $e_{23}$, $e_{24}$, $e_{35}$ and $e_{45}$ accumulate zero risk cost, since these edges lie outside of the penalty region.

Using the proposed lexicographic search, the path $\{e_{12},e_{24},e_{45},e_{56}\}$ is returned as the optimal path from $x_1$ to $x_6$, offering minimum threat exposure, and also minimum distance traveled among the competing solutions that offer identically minimal threat exposure (we note that path $\{e_{12},e_{23},e_{35},e_{56}\}$ offers identical threat exposure, but a longer distance). In the examples to follow, we explore larger cost hierarchies of up to $K=4$ in size.


\section{Algorithm Complexity}

The proposed lexicographic search, per the pseudo-code provided in Algorithm 1, takes on worst-case complexity $O(K|V|^2)$. For clarity and illustrative purposes, we have used a naive $O(|V|^2)$ implementation of Dijkstra's algorithm, describing the lexicographic search using a basic queue that could be implemented using a linked list or similar. In the worst case, (1) finding the node(s) in the queue with the minimum label (Line 9, costing $O(|V|^2)$ over the duration of the standard algorithm), and (2) expanding a node and inspecting its adjacent neighbors (Line 14, costing $O(|E|)$ over the duration of the standard algorithm) will each be repeated $K$ times, once for each cost function in the hierarchy, during every execution of the while loop. 

In the most time-efficient known implementation of Dijkstra's algorithm, which uses Fibonacci heaps \cite{fredman1987}, the complexity of the standard, single-objective algorithm is reduced from $O(|V|^2)$  to $O(|V|log|V| + |E|)$. Finding the minimum label in the graph is trivial due to the maintenance of a priority queue, but deleting a node from the heap is a $O(log|V|)$ operation that must be repeated $|V|$ times over the duration of the algorithm. Expanding a node and inspecting its adjacent neighbors continues to cost $O(|E|)$ over the duration of the algorithm, since a worst case of $O(|E|)$ label updates must be performed in the heap, each of which costs $O(1)$. To adapt this to a lexicographic search, the nodes in the heap must be prioritized per the lexicographic ordering of the graph nodes, so that the minimum label reflects not only the minimum primary cost, but the optimum according to the formulation given in (1). Although only one node will be deleted from the heap in each iteration of the algorithm's while loop, each of the $O(log|V|)$ comparisons required will take $O(K)$ time, and so the cost of node deletion over the duration of the algorithm will increase to $O(K|V|log|V|)$.

\begin{figure*}[ht]
\centering
\subfigure[Example 1]{\includegraphics[width=.3\textwidth]{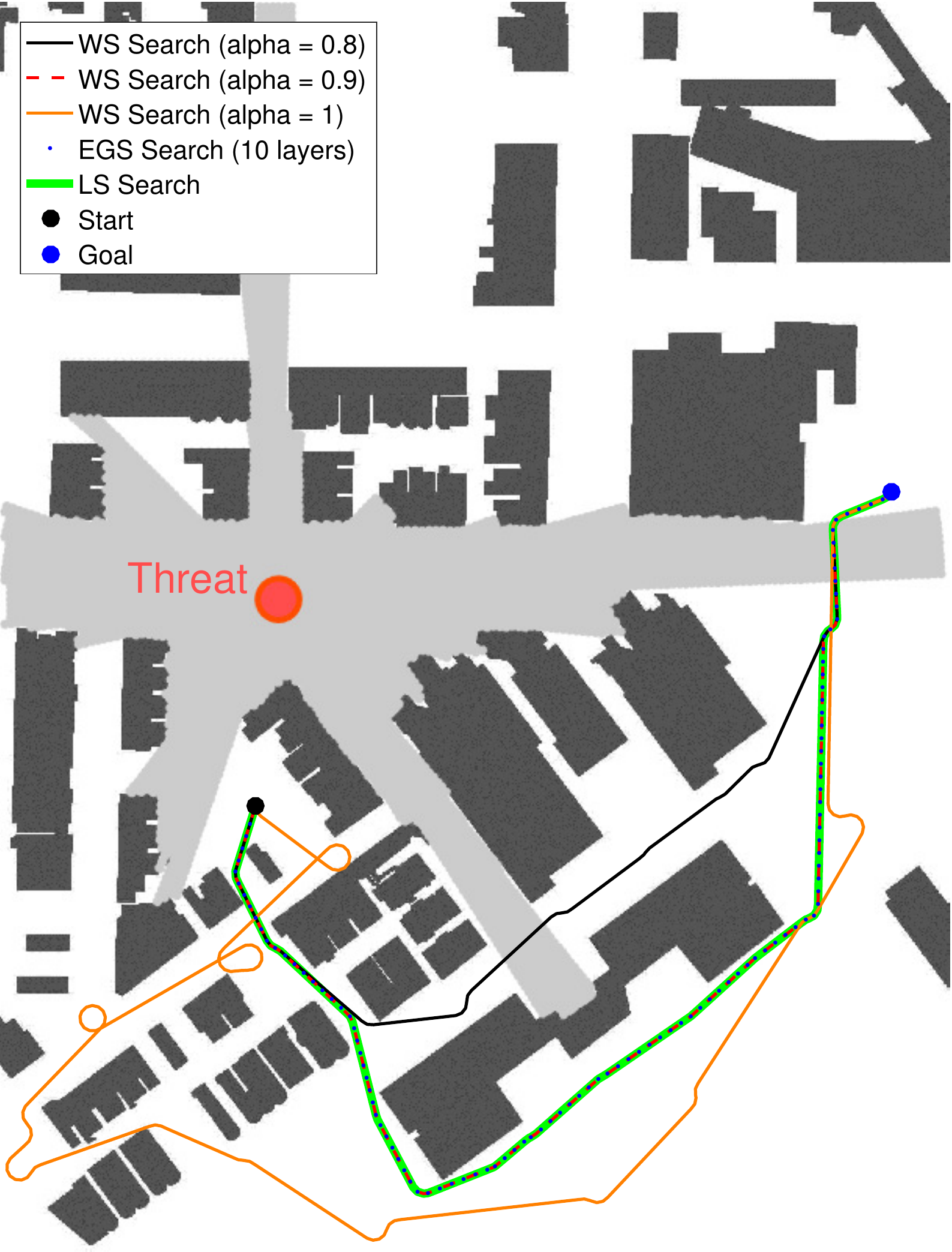}}
\hspace{3mm}
\subfigure[Example 2]{\includegraphics[width=.3\textwidth]{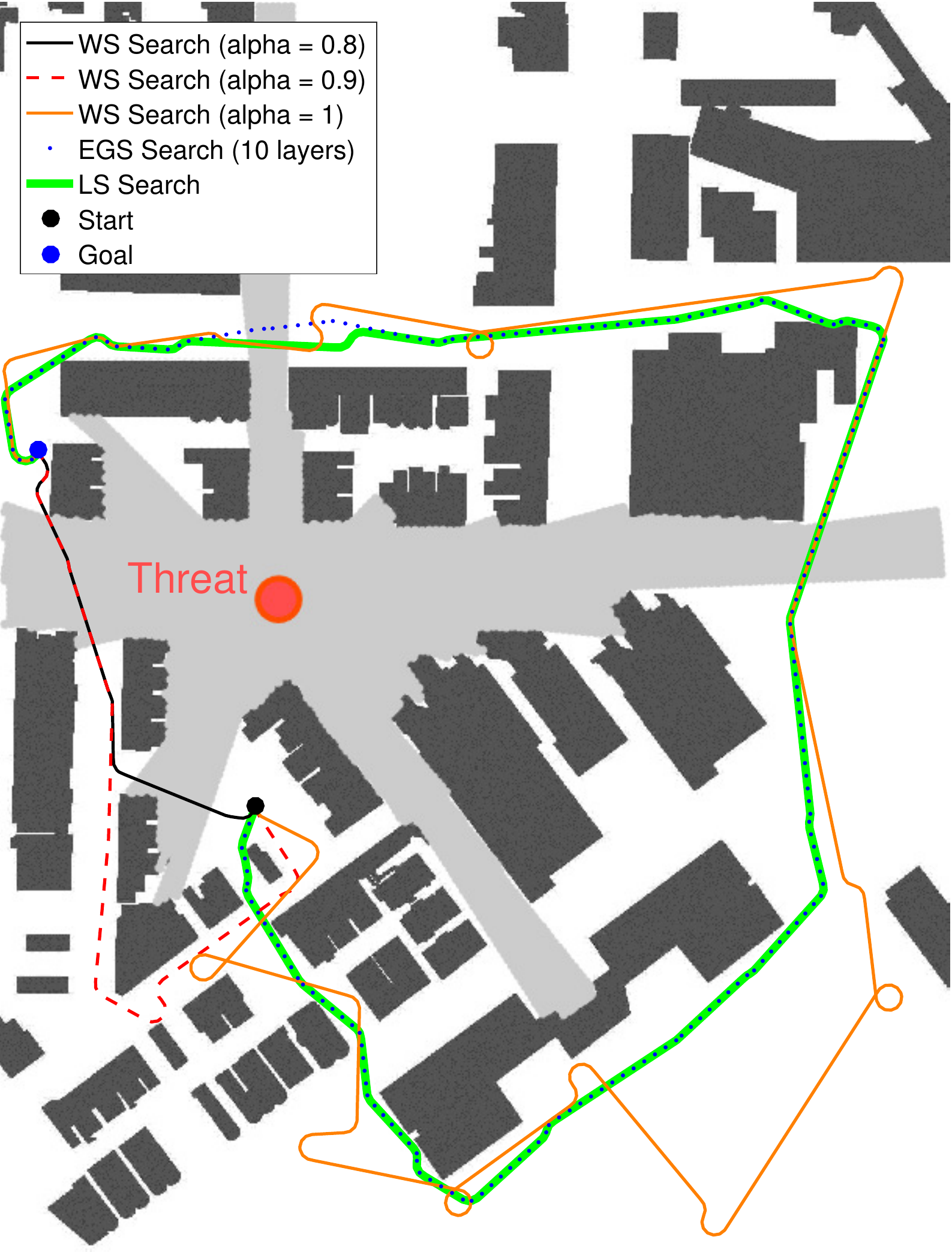}}
\hspace{3mm}
\subfigure[Example 3]{\includegraphics[width=.3\textwidth]{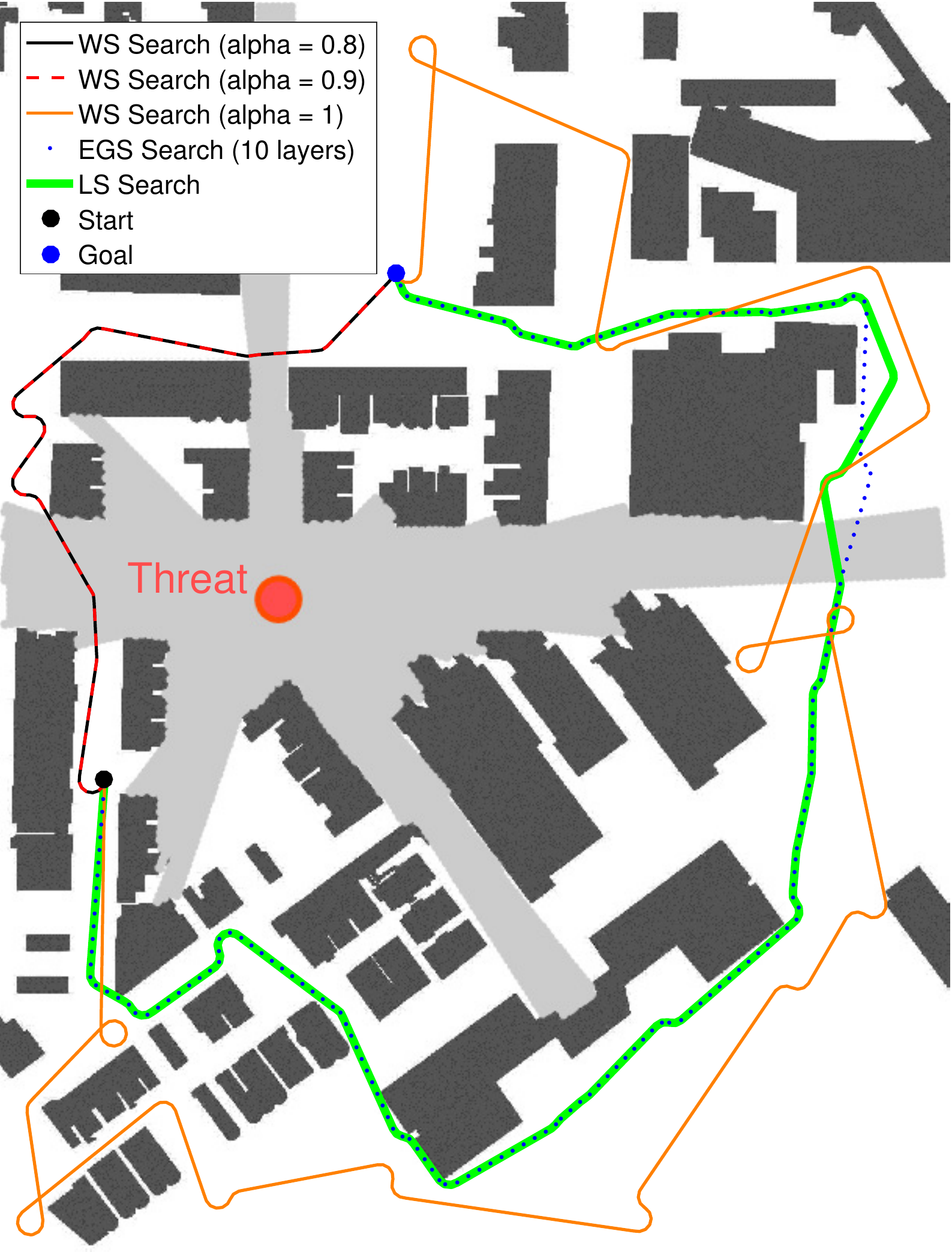}}
\caption{The solutions generated by three competing methods are plotted in different colors. The red dot indicates a threat, and the regions visible to the threat are colored gray. Obstacles are colored black. Budgets are discretized evenly among 10 layers for the EGS method. 2000 nodes are included in the roadmap used in each example. Note that LS, WS($\alpha=0.9$) and EGS can find the same optimal solution in (a).}
\label{fig:bi-path}
\vspace{-0mm}
\end{figure*}

The costs associated with the alteration of labels in the heap will also reflect the $K$ cost functions being considered. To maintain a lexicographic ordering among the nodes in the heap, all nodes undergoing label changes during an iteration of the algorithm's while loop may have their labels individually adjusted as many as $K$ times. Akin to the steps performed in lines 15-20 of Algorithm 1, this is the worst-case number of times a node's label must be adjusted to establish the correct lexicographic ordering among a set of nodes with $K$ cost functions. Over the duration of the algorithm, this will result in a worst-case $O(K|E|)$ label changes within the heap, each of which carries $O(1)$ complexity. As a result, the worst-case complexity of a lexicographic search using a Fibonacci heap will be improved to $O(K|V|log|V| + K|E|)$, from the original $O(K|V|^2)$. In the results to follow, we opt to implement and study the $O(K|V|^2)$ version of the algorithm in software, due to its ease of implementation and efficient memory consumption.

We also note briefly that an adaptation of Dijkstra's algorithm is selected in this application due to the fact that all graphs considered are characterized by non-negative, time-invariant edge weights. The consideration of negative edge weights would require an adaptation of the Bellman-Ford or Floyd-Warshall algorithm \cite{clrs}, and the consideration of time-varying weights, such as those which might depend on the action or measurement history of a robot, as frequently occurs in belief space planning, may require search algorithms of exponential complexity \cite{brm}.



\section{Experimental Results}

\subsection{Bi-criteria Planning for a Dubins Vehicle}

To permit a fair evaluation of the proposed algorithm's computational performance, we must select competing algorithms that form a suitable basis for comparison. Our first basis for comparison involves testing different combinations of multiplicative weights on the competing cost functions. The weighted terms are summed together into a single composite cost function, and used to generate edge costs that are subjected to a standard Dikjstra search over a roadmap. This method, employing a weighted sum of the competing costs, is abbreviated as WS in the discussion to follow, and its cost function $f_{WS}(\sigma)$ is given in Equation (2) for a bi-criteria problem.
\begin{align}
&f_{WS}(\sigma)=\alpha f_1(\sigma) + (1-\alpha)f_2(\sigma), \;\;\; \alpha \in (0,1) 
\end{align}
Another basis for comparison is a multi-objective search method proposed for graphs over which costs are non-negative, but not strictly positive (we refer the reader to the description from \cite{takei2011}, Remark 2.1(3)). This approach entails the search of an \textit{expanded graph}, which applies one of the competing resources to its edge weights, and subsequent resources are represented by expanding the original graph across many separate layers, representing different amounts of resource consumption. We hereafter refer to this competing method as the expanded graph search (EGS), which requires a layer-by-layer Dijkstra search to uncover solutions that trade off consumption of the competing resources. We have encountered the best results by placing a bi-criteria problem's secondary cost, which is strictly positive (typically a robot's distance traveled), along the edges of the roadmap. The bi-criteria problem's primary cost, which is occasionally zero-valued, accumulates from layer to layer of the expanded graph. At every layer of the graph, the Dijkstra search is constrained by its ``budget'' (the primary cost for bi-criteria problem), which cannot be exceeded by the paths recovered on this layer. 


\begin{figure}[th!]
\centering
\subfigure[Construction time]{\includegraphics[width=.49\columnwidth]{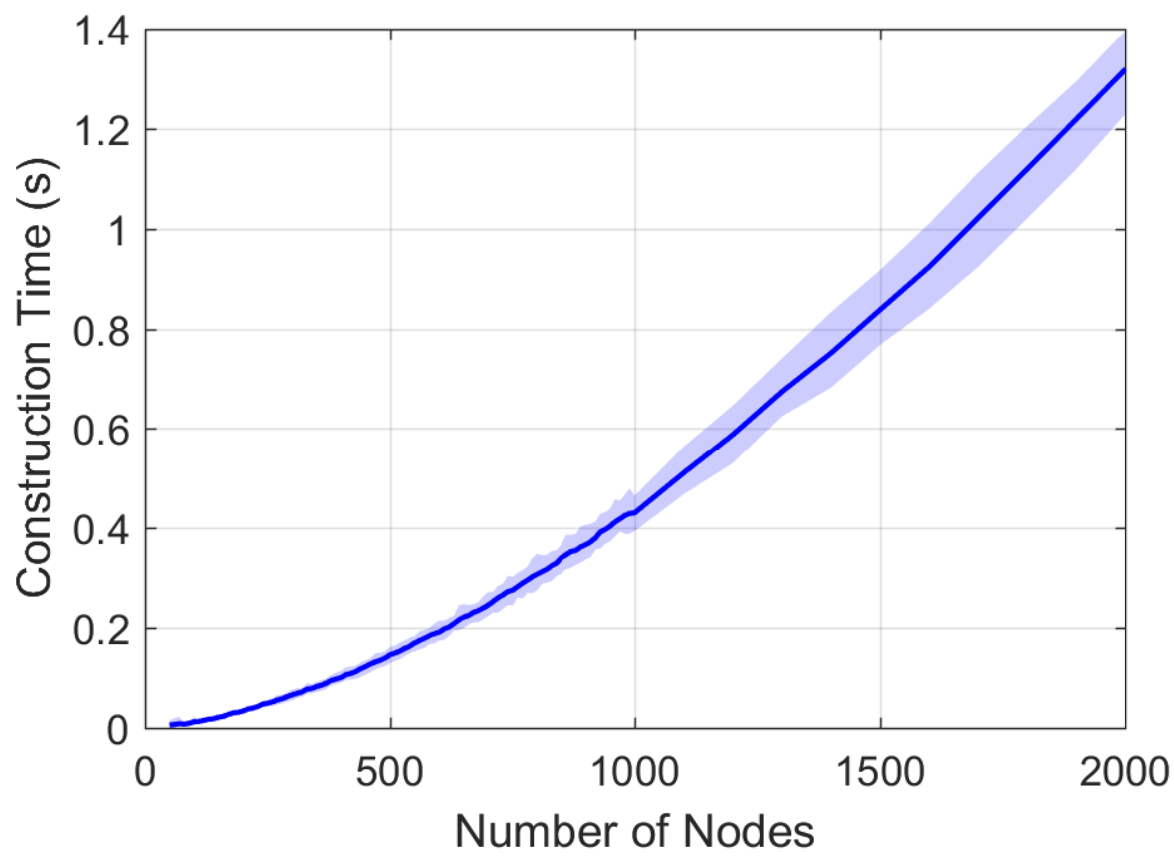}}
\subfigure[Search time]{\includegraphics[width=.49\columnwidth]{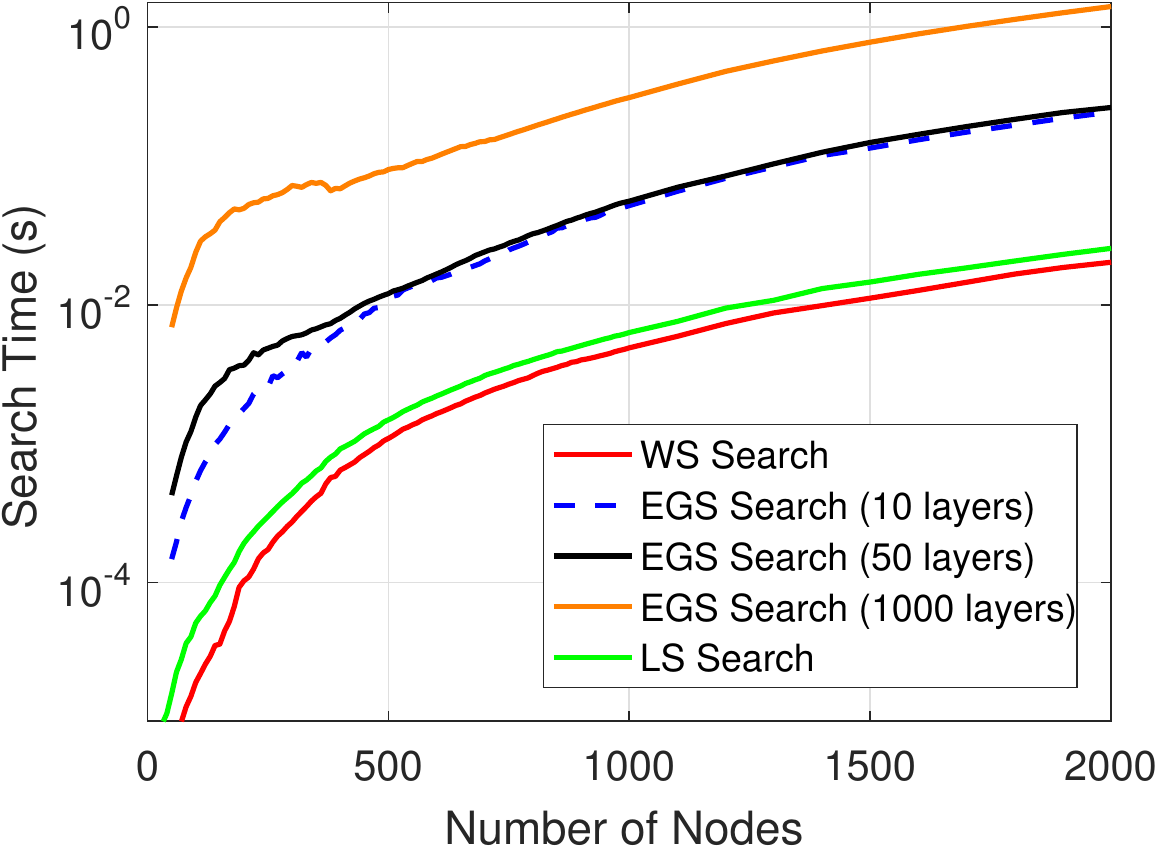}}
\subfigure[10 layers]{\includegraphics[width=.49\columnwidth]{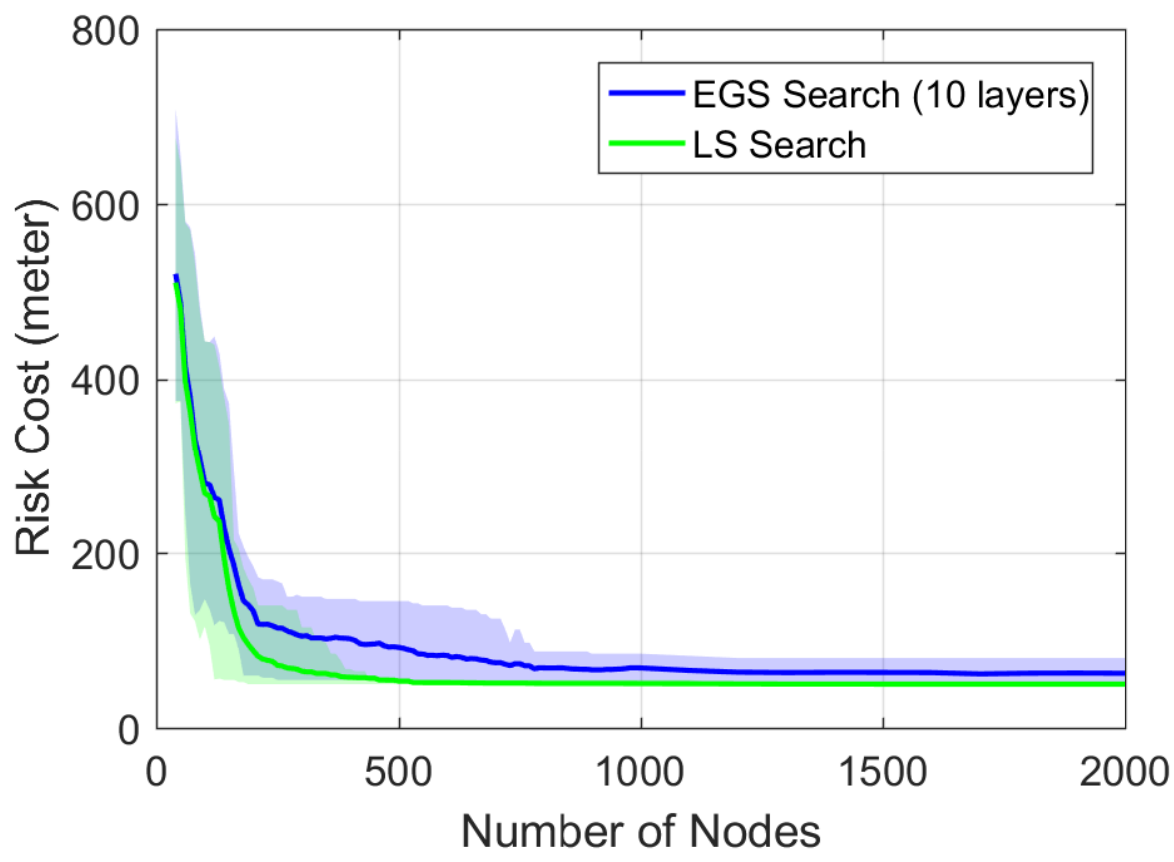}}
\subfigure[50 layers]{\includegraphics[width=.49\columnwidth]{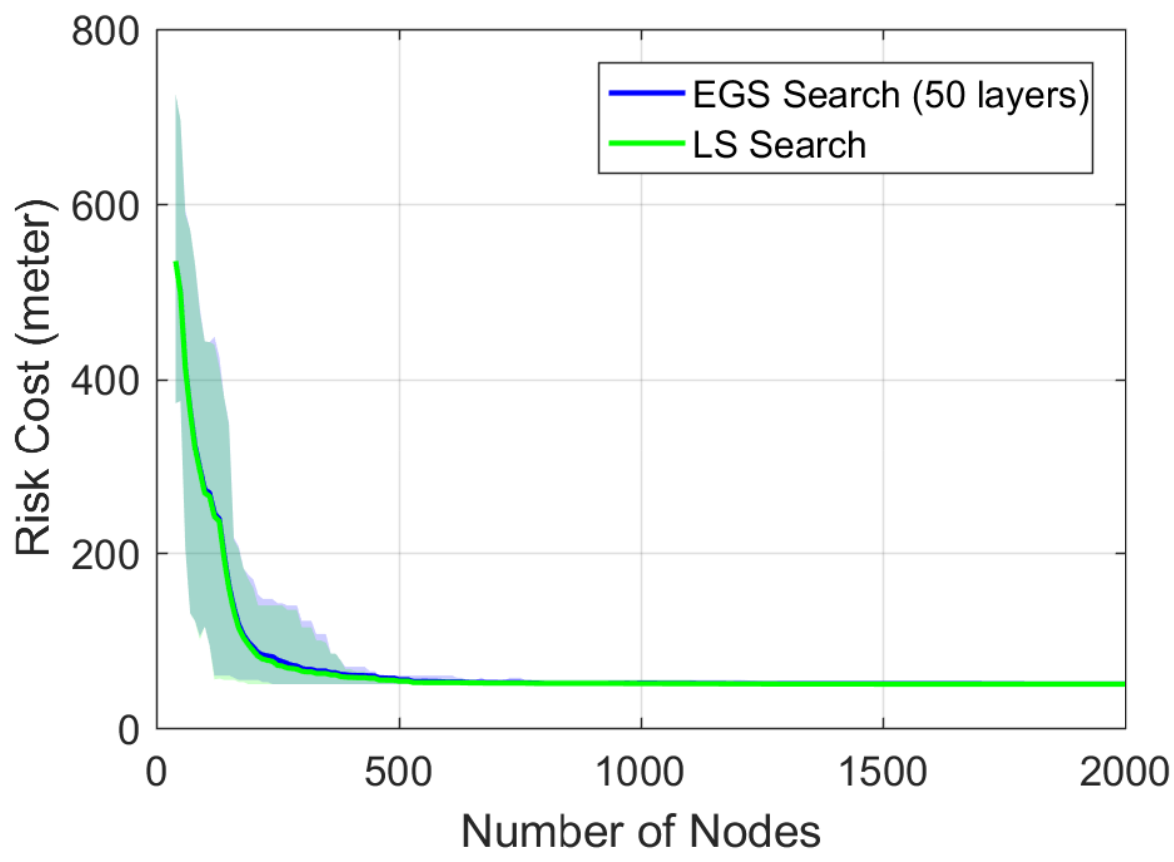}}
\subfigure[10 layers]{\includegraphics[width=.49\columnwidth]{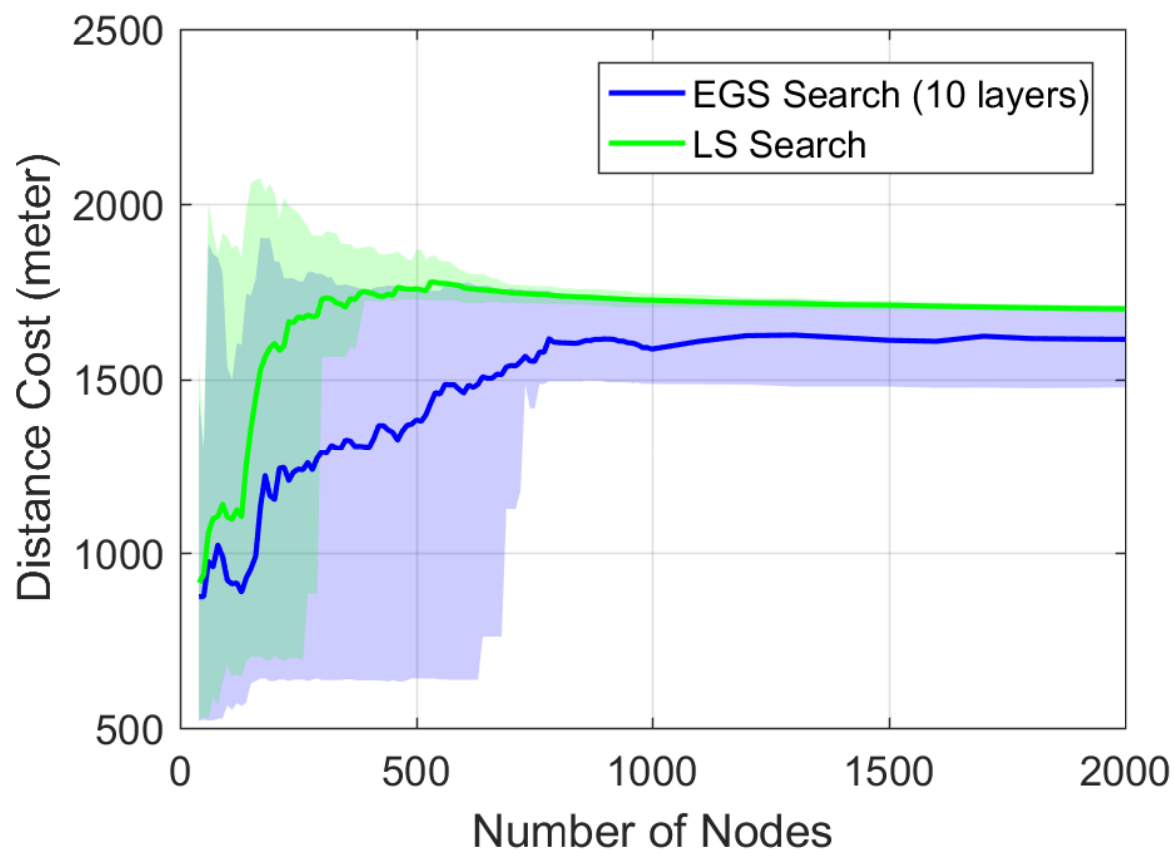}}
\subfigure[50 layers]{\includegraphics[width=.49\columnwidth]{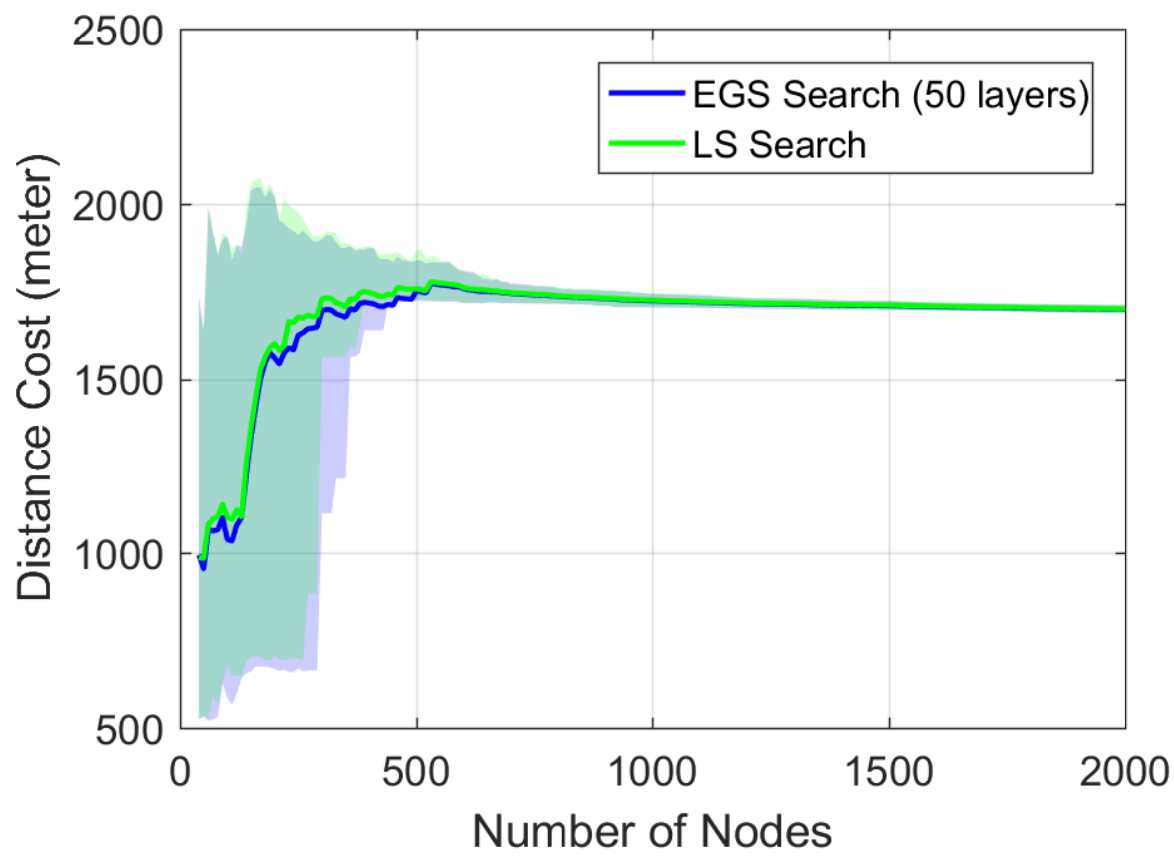}}
\caption{Roadmap construction time, search time, threat exposure cost and path duration are plotted as a function of the number of PRM* nodes, for example 3 in Figure \ref{fig:bi-path}(c), averaged over 50 trials. The mean value is plotted as a solid line, and the shaded regions indicate the 10th-to-90th percentiles.}
\label{fig:two-costs}
\vspace{-0mm}
\end{figure}

A comparison is employed to verify the effectiveness of motion planning with the proposed lexicographic search (LS) method. For the purposes of illustration, our first example considers planning in $SE2$ for a constant-velocity Dubins vehicle, in a bi-criteria problem that considers the tradeoff between a robot's travel distance and threat exposure. The black and blue points in Figure \ref{fig:bi-path} indicate the robot\textquoteright s initial and goal locations respectively. The robot is within line-of-sight to a hostile threat in regions of the workspace that are colored gray. For the LS framework, the primary, threat-exposure cost is the distance traveled in the gray regions, and the secondary cost is the total distance traveled from $x_{init}$. 

The three competing search methods are compared over the same PRM* \cite{karaman2011} graph in each of 50 trials for each example, performed across different weights $\alpha$ and ``budget'' choices. We assign three different values for $\alpha$: 0.8, 0.9 and 1. The maximum allowable risk budget for the EGS method is set to 500 meters of exposure. The risk budget is discretized evenly from 0 to 500 divided among 10, 50 and 1000 ``budget layers''. For simplicity, budget layers for EGS are referred as ``layers" in the comparisons to follow.

Three representative examples with different start and goal locations are shown in Figure \ref{fig:bi-path}. In Figure \ref{fig:bi-path}(a), the path of WS ($\alpha = 0.9$), which has the lowest risk cost, is the same as the optimal path from LS. When $\alpha=0.8$, the path of WS traverses more in the visible regions and is not optimal. Then in Figure \ref{fig:bi-path}(b), WS ($\alpha=0.9$) fails to recover the optimal path that LS can. However, the dashed red path($\alpha=0.8$) has a smaller risk cost than the black path($\alpha=0.8$). In Figure \ref{fig:bi-path}(c), WS, when $\alpha = 0.8$ and $\alpha = 0.9$, fails to recover the lowest risk cost path both times. Note that WS can always find the lowest risk cost solutions by setting $\alpha$ to 1, however there would be \textit{no} weight to influence the formation of paths outside the gray region, and ties would be broken arbitrarily.

After implementing 50 trials of comparison for each example, WS has been observed to be \textit{highly sensitive} to the choices of $\alpha$ and the initial and goal positions, where very small differences can have a large impact on the quality of the solutions obtained. A systematic procedure to use such a method would involve choosing many values of $\alpha$, searching the roadmap with each, and adopting values whose paths offer minimum composite cost. However, a straightforward linear scaling of $\alpha$ has not been found in practice to thoroughly probe the space of solutions, without being exhaustive. Meanwhile, a thorough search over different $\alpha$ values is computationally more expensive than a single lexicographic search. We also notice that WS fails to recover optimal paths for some start and goal locations even when we are using a systematic procedure to find $\alpha$.

EGS, which outperforms WS, as it can achieve optimality with finely discretized layers, is a systematic approach for multi-objective planning that adopts a secondary budget as a constraint. However, EGS is also observed to be sensitive to the number of layers and the initial and goal locations (Figure \ref{fig:bi-path}(a) and (b)). As is shown in Figure \ref{fig:two-costs}(c) and (d) for example 3, in order for the paths of EGS to match the levels of threat exposure recovered from the LS method, a significantly greater computational investment is required. However, the LS method does produce longer paths, due to the primary objective of minimizing the robot's cumulative exposure to threat. As EGS layers are discretized more finely, the paths produced by EGS gradually approximate the solution from the LS method. We notice that EGS can recover the same paths that are returned by LS when the budget for EGS is discretized into 1000 layers (only search time is shown in Figure \ref{fig:two-costs}(b)). However, such number of layers cannot guarantee the optimality of a path when the initial and goal positions are changed. On the contrary, the optimal solutions can always be found by LS.

Using C++ implementations of the three algorithms run on a laptop with an Intel i7 2.5GHz processor, equipped with 16GB RAM, algorithm runtime results of example 3 are shown in Figures \ref{fig:two-costs}(a) and (b). Note that only one WS runtime is shown. If we try different choices of $\alpha$ more than once, WS takes more time than LS. Obviously, the LS method offers a substantial computational advantage over the EGS method. Since discretization of budget is required by EGS, the roadmap is searched repeatedly on the many resulting layers that represent different values of budget. If the budget is discretized finely enough, EGS can recover risk-optimal paths for a given start and goal location. However, if the budget is not discretized finely enough, we are left to guess where finer discretization is needed, and we must reason in unintuitive units. 

\begin{figure}[h!!]
\centering
\subfigure[]{\includegraphics[width=.95\columnwidth]{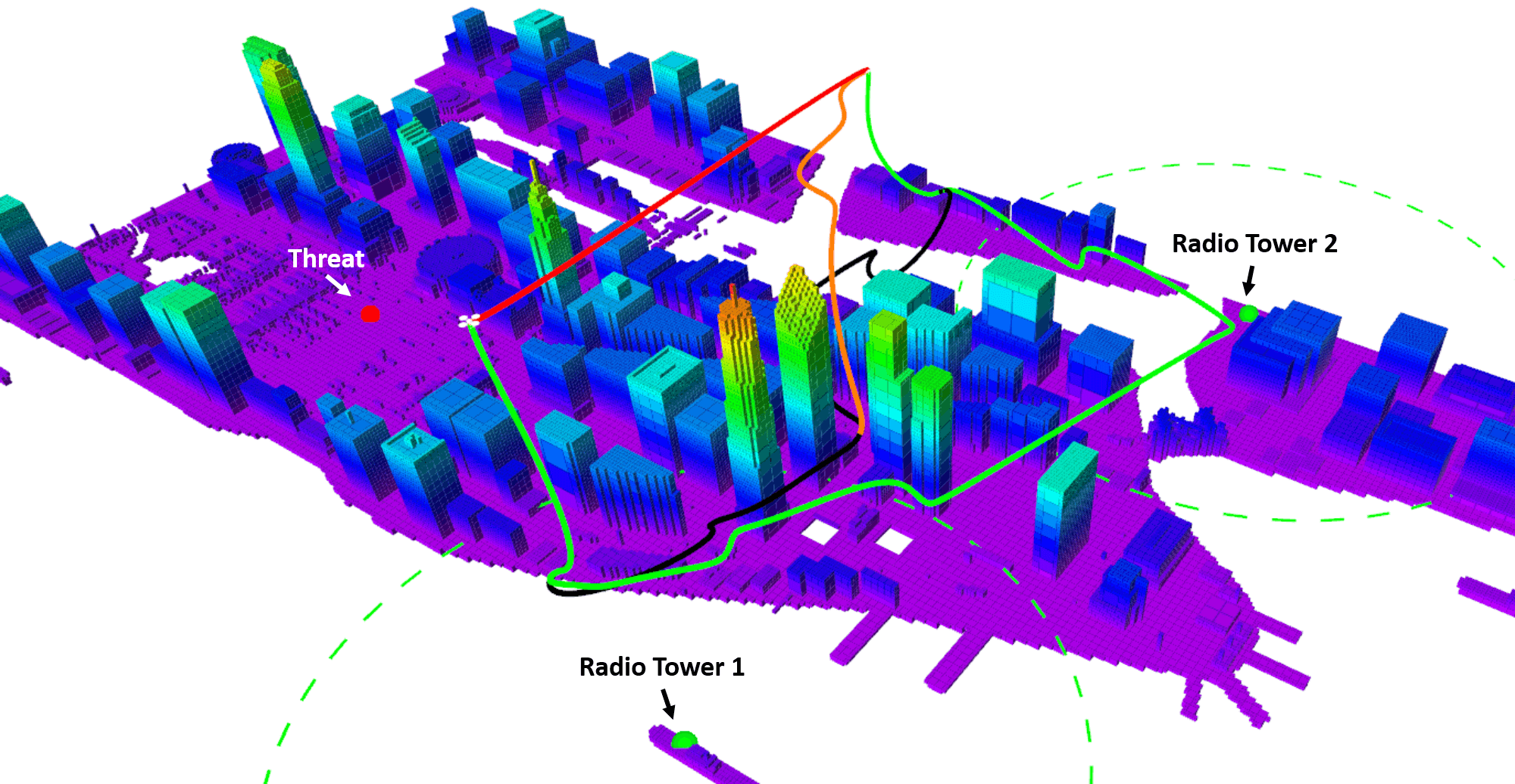}}
\subfigure[]{\includegraphics[width=.475\columnwidth]{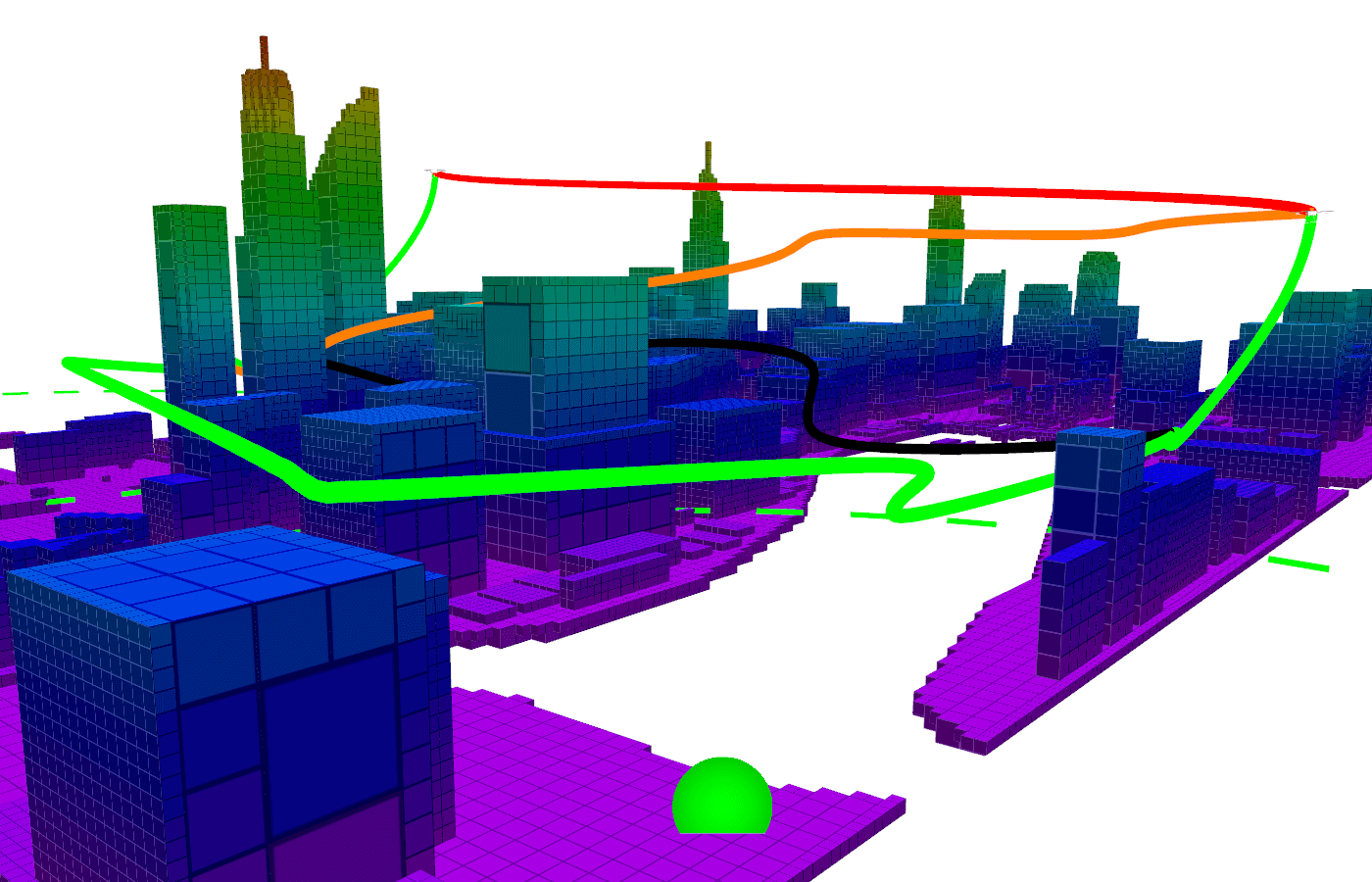}}
\subfigure[]{\includegraphics[width=.475\columnwidth]{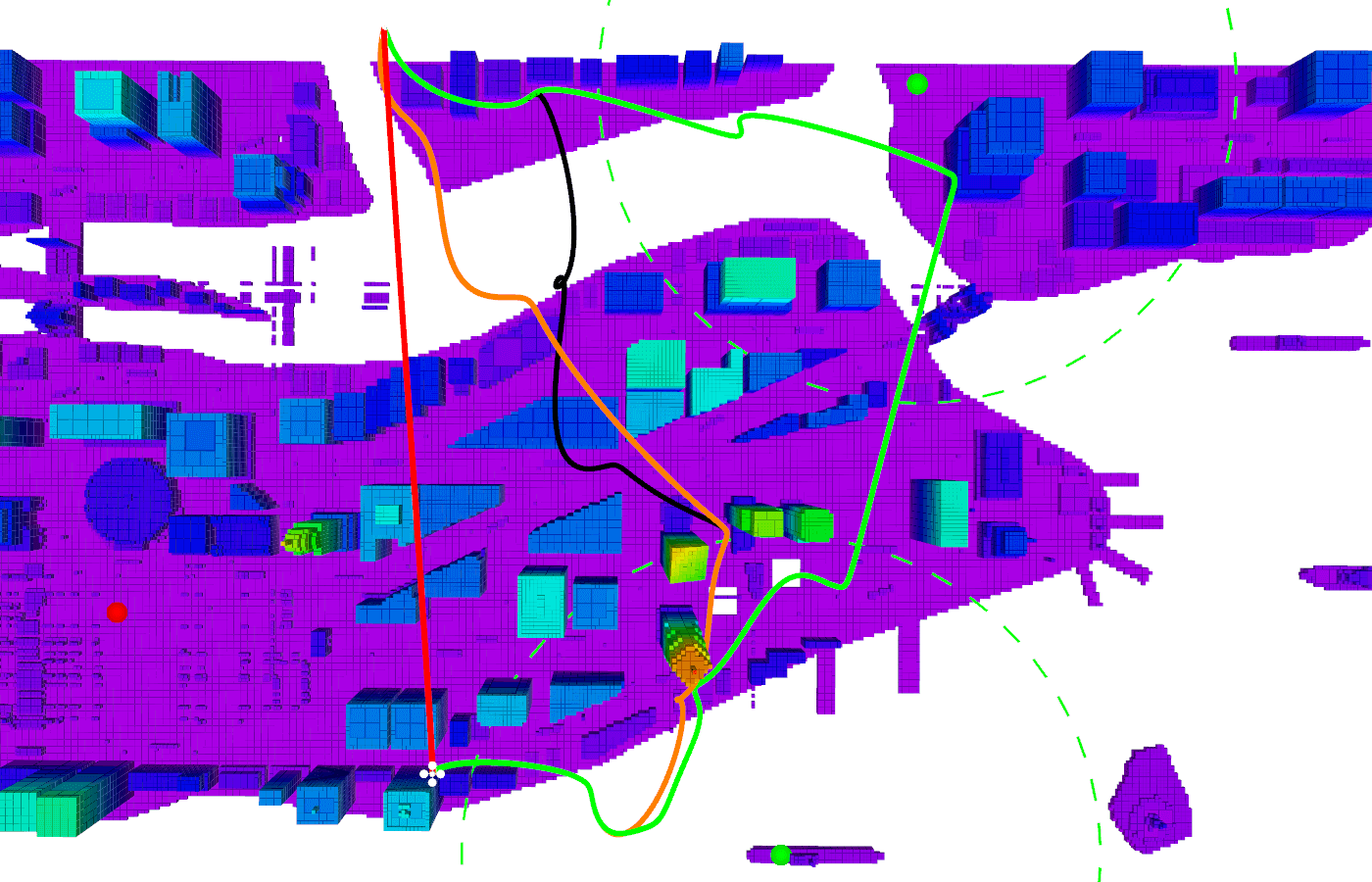}}
\caption{LS is applied to a quad-criteria UAV planning problem. The color of buildings in the scenario changes from purple to yellow as altitude increases. The paths that are colored red, orange, black and green are returned when applying a D criterion, R-D criteria, R-L-D criteria and R-L-C-D criteria respectively. A 5000 node-roadmap is used in this example. The UAV is enlarged for visualization. }
\label{fig:uav-path}
\vspace{-0mm}
\end{figure}

In addition, the maximum budget for EGS must be selected by the user. On the one hand, if the maximum budget is too large, the budget needs to be discretized more finely, which will cause unnecessary computational burden. Meanwhile, EGS prefers shorter paths when the budget headroom is available, and the algorithm will likely fail to recover some paths achieving low threat exposure. On the other hand, paths that successfully reach the goal may not be recovered if the maximum budget is assigned too small a value. Overall, the lexicographic search method demonstrates efficiency and optimality for multi-objective planning without the need for parameter-tuning.

\subsection{Quad-criteria Kinodynamic Planning for a Quadrotor}

We next explore a kinodynamic UAV motion planning problem, in which a quadrotor model \cite{vdb}, which is 10-dimensional, is linearized about the aircraft's hover point. Its state can be expressed as $x = (p,v,r,w)^{T}$, where $p$ and $v$ are three-dimensional position and velocity, $r$ and $w$ are two-dimensional orientation and angular velocity, and yaw and its derivative are constrained to zero after linearization. We assume there is a threat in the workspace that can place the UAV in danger if it is seen. A laser rangefinder, which has a 30 meter range, is mounted downward under the UAV and is used for improved-precision localization via scan-matching. There are also two ultra-wide band radio towers that the UAV can communicate with to upload mission data when it is within range.

\begin{figure}[ht]
\centering
\subfigure[Roadmap construction time]{\includegraphics[width=.49\columnwidth]{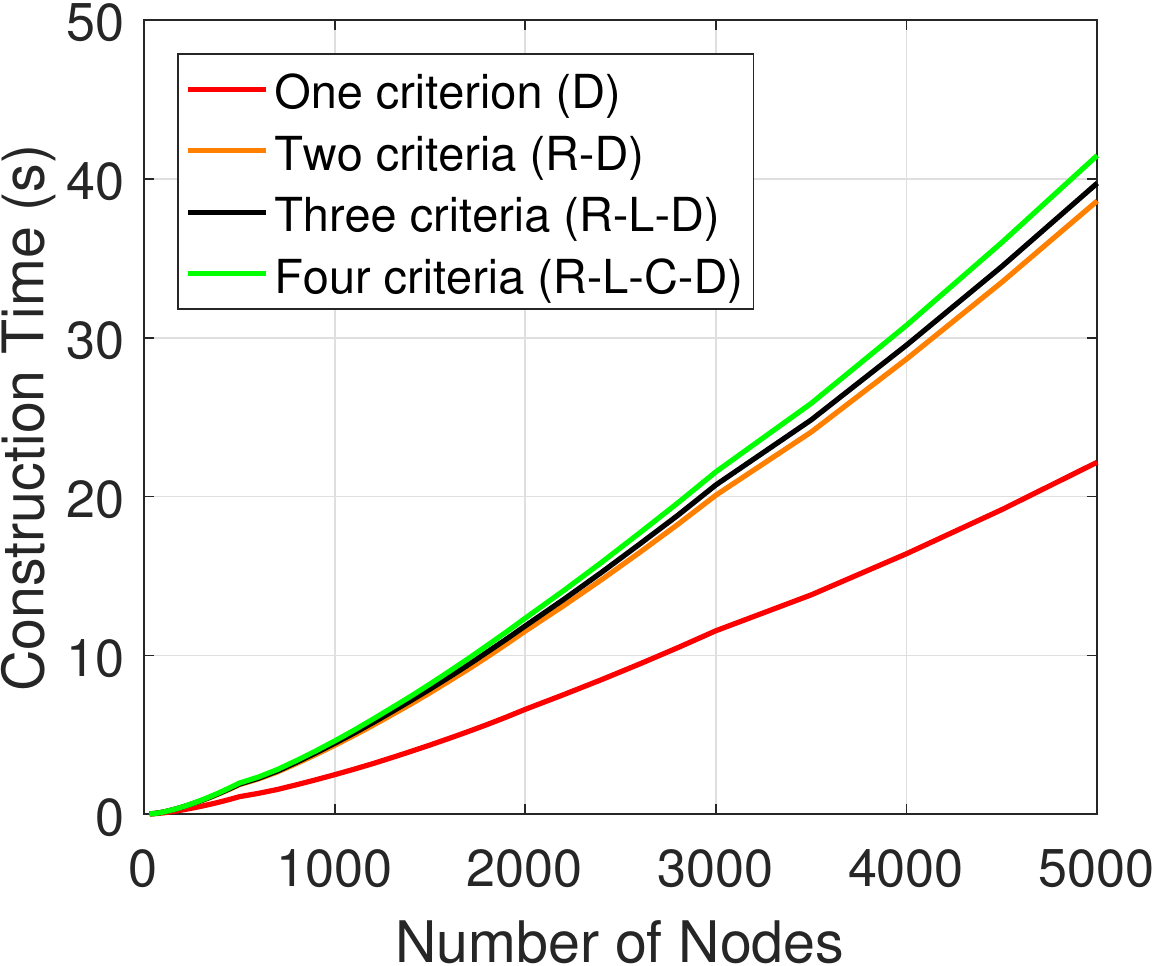}}
\subfigure[Lexicographic search time]{\includegraphics[width=.49\columnwidth]{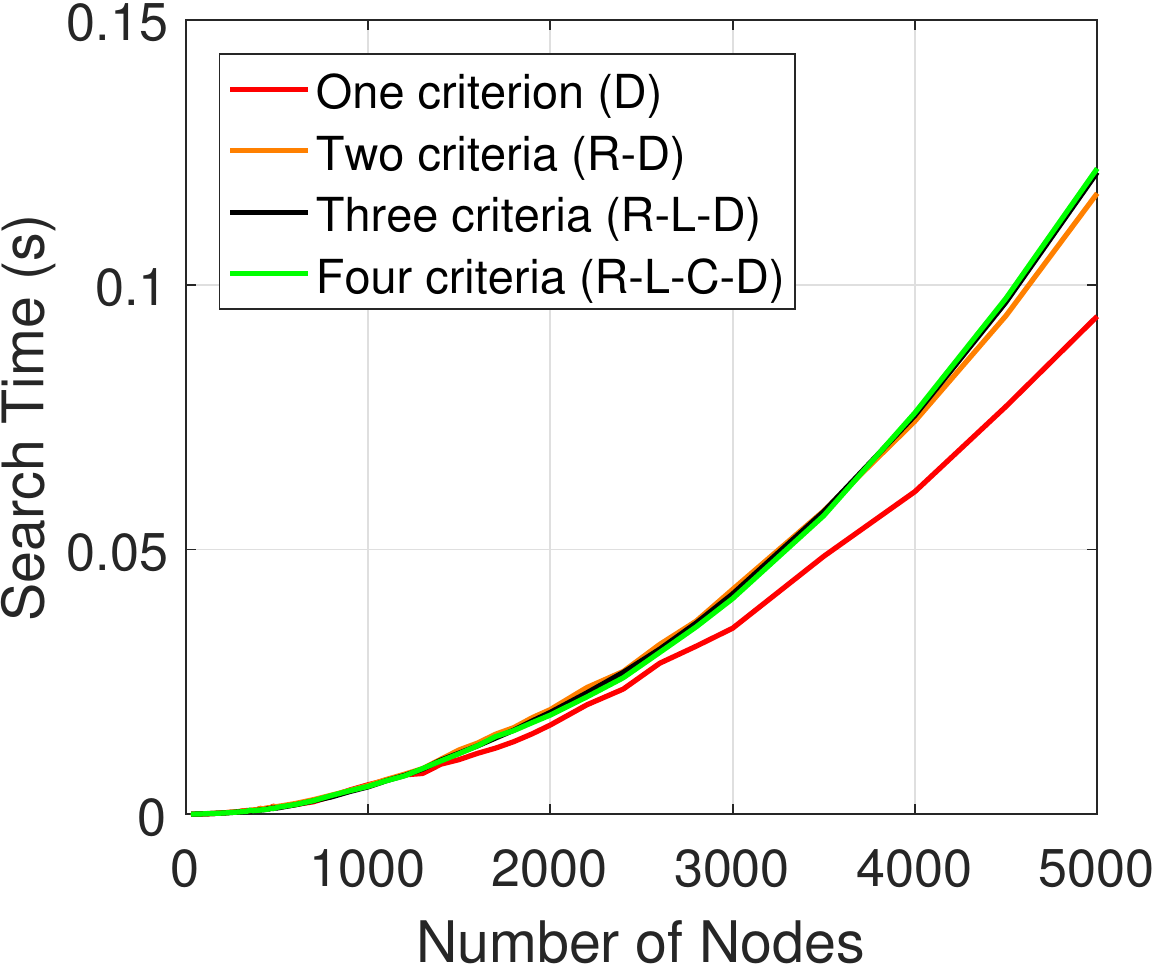}}
\caption{Roadmap construction time and search time for UAV motion planning, as a funtion of the number of nodes, with different cost criteria combinations, averaged over 50 trials.}
\label{fig:uav-time}
\vspace{-0mm}
\end{figure}

\begin{figure}[ht]
\centering
\subfigure[Risk cost]{\includegraphics[width=.49\columnwidth]{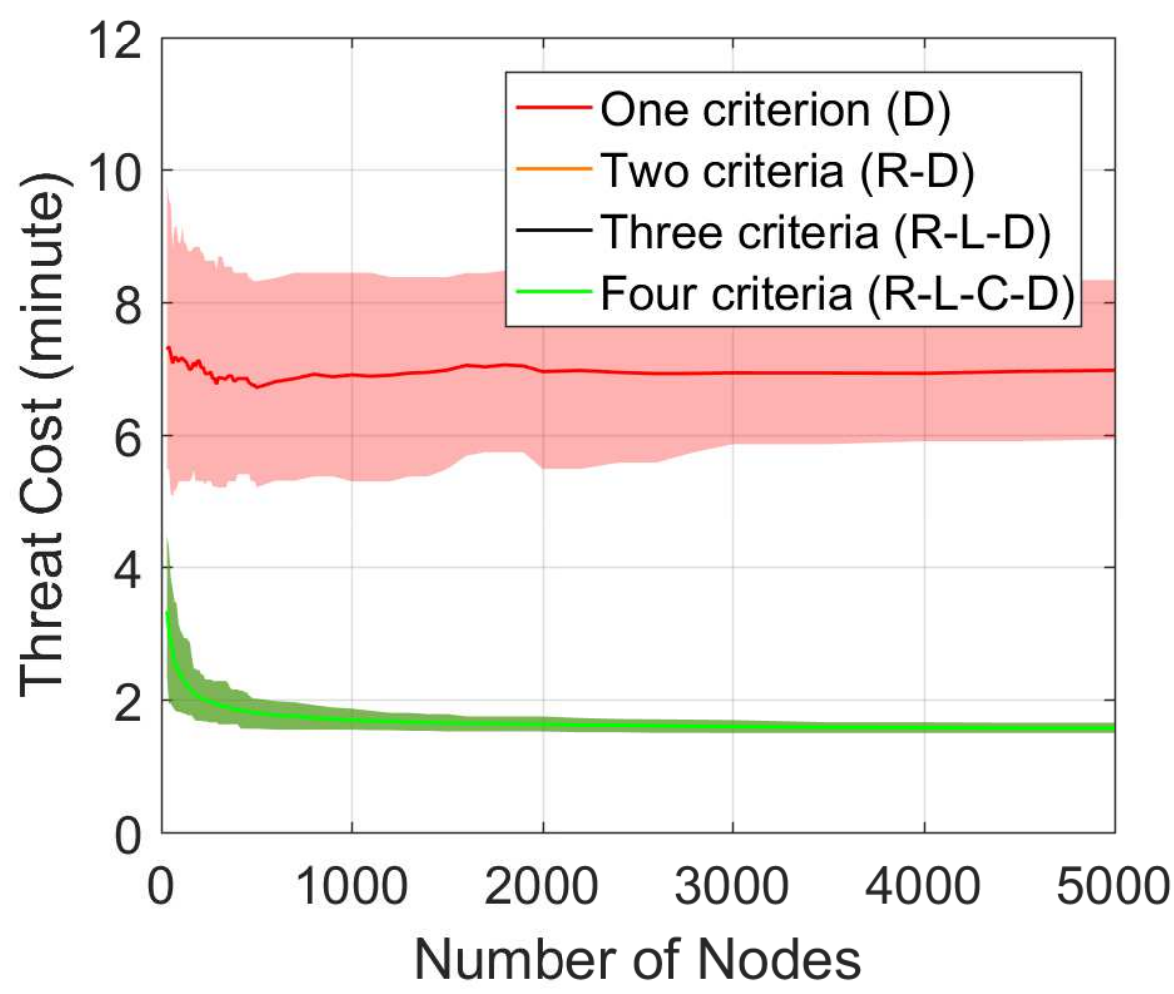}}
\subfigure[Localization cost]{\includegraphics[width=.49\columnwidth]{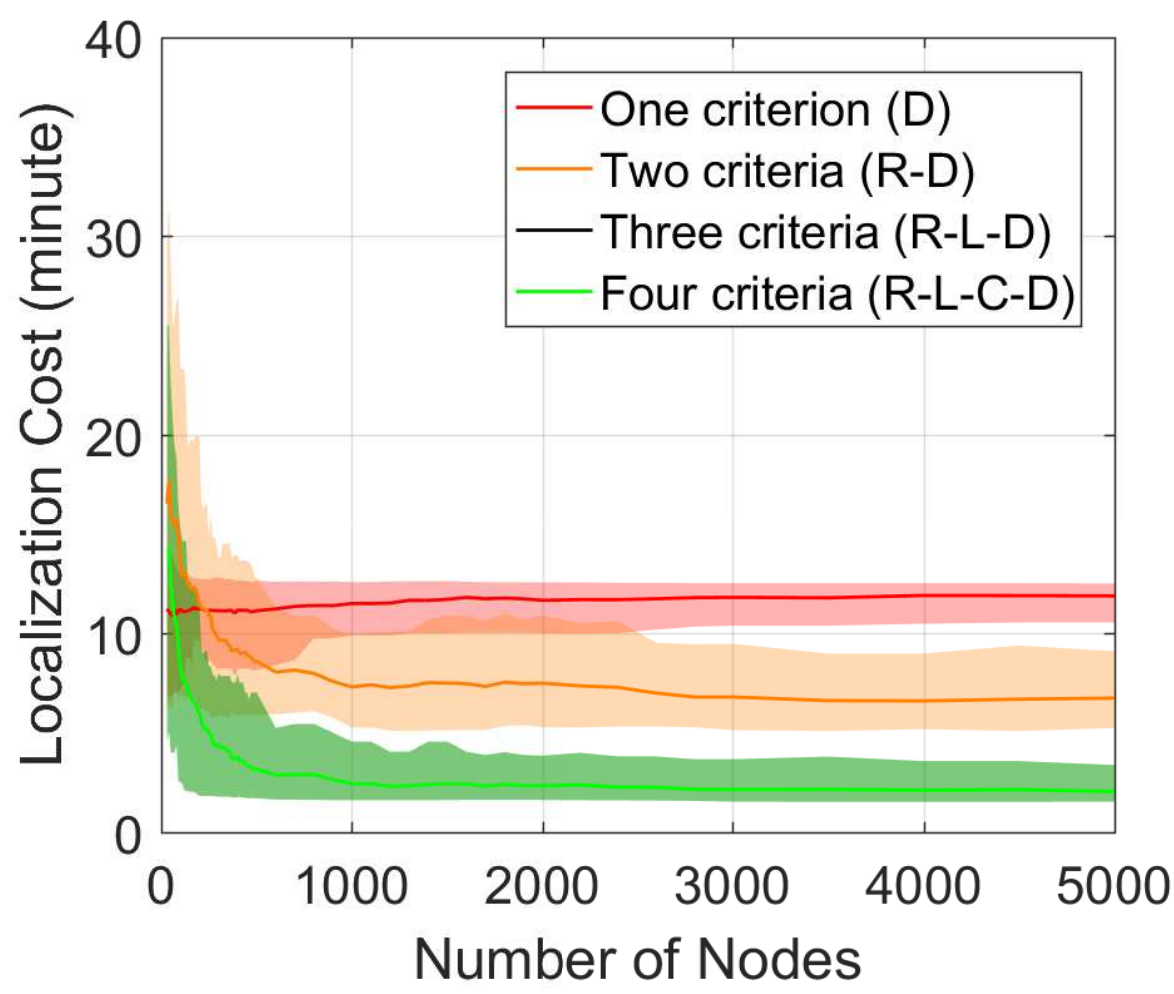}}
\subfigure[Communication cost]{\includegraphics[width=.49\columnwidth]{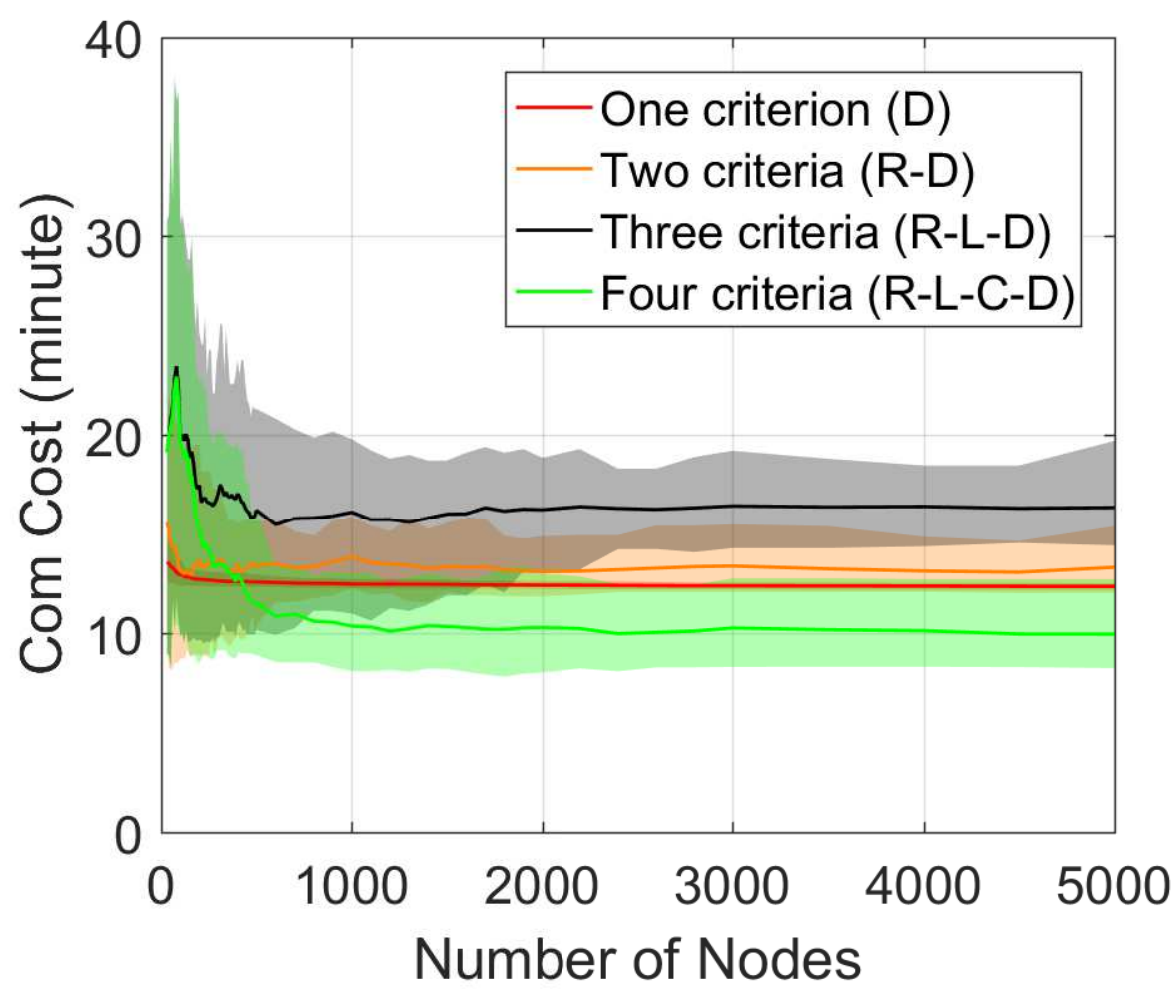}}
\subfigure[Distance cost]{\includegraphics[width=.49\columnwidth]{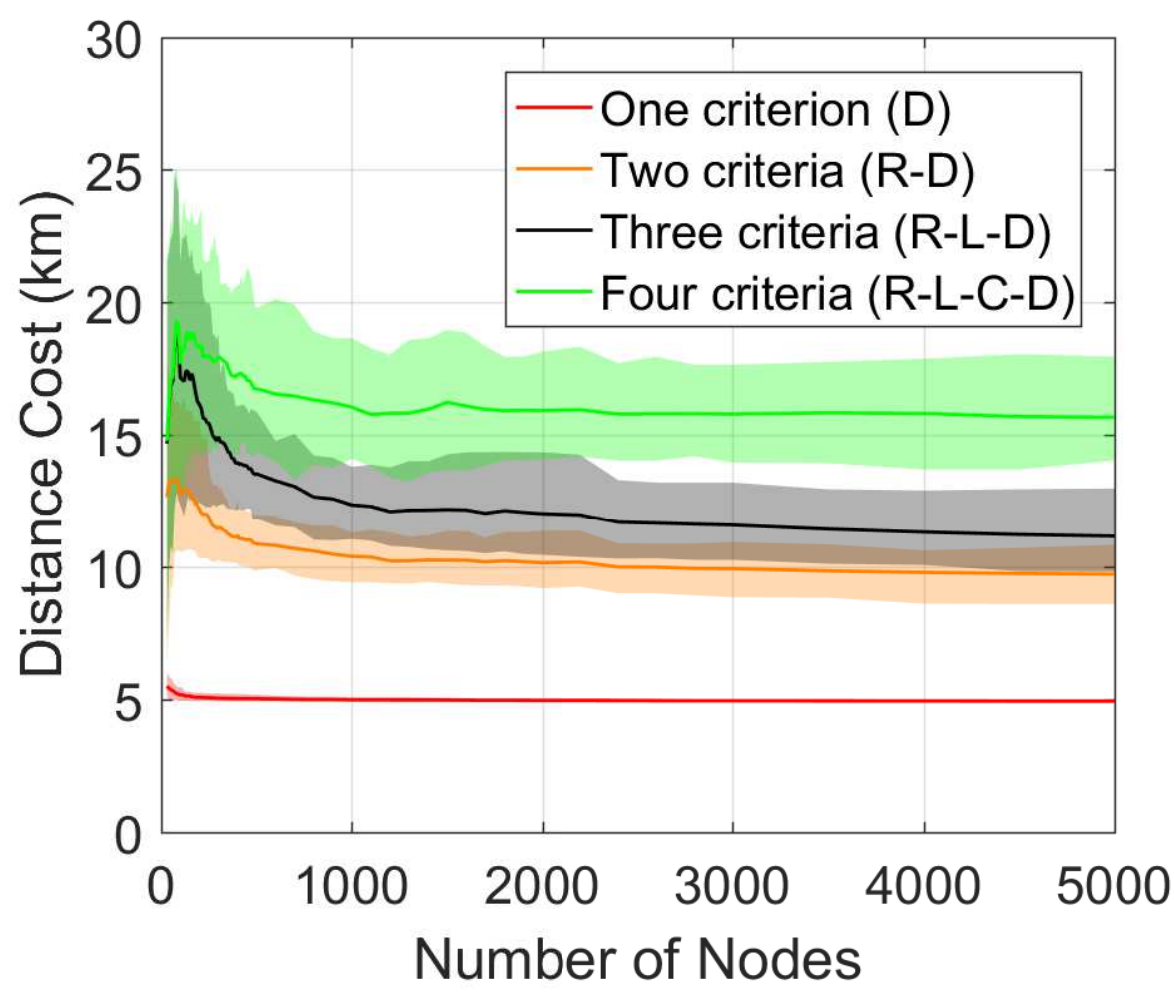}}
\caption{The results of LS with four cost criteria combinations are plotted as a function of the number of nodes in the PRM* roadmap, averaged over 50 trials. Mean value is plotted as a solid line, and the shaded regions indicate the 10th-to-90th percentiles.}
\label{fig:uav-cost}
\vspace{-0mm}
\end{figure}

In order to explore the influence of different lexicographic orderings on the UAV planning solution, four types of cost criteria are considered in this example. These criteria, listed in descending order of importance, are threat exposure, localization cost, communication cost and distance traveled. Threat exposure represents the time duration in which the UAV is visible to the threat. When the UAV cannot be seen by the threat, this cost is zero-valued. Localization cost is the time duration for which no buildings or land features are within the sensing range of the LIDAR. Similarly, communication cost is the time duration for which radio tower reception is out of range. The final cost criterion is distance traveled, which is strictly positive. The WS and EGS methods are not employed here due to computational intractability when faced with so many cost criteria. As more cost criteria are imposed in a planning problem, we need to fine-tune weights based on unintuitive units. The systematic way to find the best weight combination becomes intractable as the number of weight combinations increases exponentially. Similarly, the EGS method's expanded graph grows prohibitively larger, with discretized budget layers in multiple dimensions.

Four combinations of cost criteria are considered: (I) distance only (D), (II) a risk-distance ordering (R-D), (III) a risk-localization-distance ordering (R-L-D) and (IV) a risk-localization-communication-distance (R-L-C-D) ordering, and an optimal trajectory is produced for each ordering. Note that all four cost functions are evaluated over each trajectory solution in order to examine the benefits of the proposed search method. The resulting trajectories are shown in Figure \ref{fig:uav-path}. When only a distance criterion is considered, the search is a standard Dijkstra search and we are given the shortest path (red). However, this path yields the highest risk cost, as the UAV is visible to a threat most of the time. When risk cost is considered, LS is needed for breaking ties when the UAV is not visible to the threat. The resulting trajectory, which is colored orange (orange, black and green lines share the same beginning portion), results in a quick dive to avoid being seen by the threat and traversal among buildings before reaching the goal. Nevertheless, the latter half of the path may lead to high localization cost since only distance is considered when the UAV is not exposed to threats. After adding a localization criterion to the cost hierarchy, the resulting trajectory, colored black, spends more time flying close to buildings and land features. Finally, the upload of data may be valuable for analysis of the UAV's performance, and so communication cost is added to the hierarchy for this purpose. Consequently, the resulting trajectory takes a detour to spend more time within range of the two radio towers for communication, while mitigating risk and localization costs. The green path shows the solution when all four criteria are used in the search.

Figure \ref{fig:uav-time} displays the time required to construct the roadmaps, and to perform the multi-criteria searches, needed to solve the UAV planning problem. Per the results in Figure \ref{fig:uav-cost} (b) (c) and (d), we note that the latter three paths (orange, black, and green) share identical risk cost, since threat exposure is assigned the highest priority in all three instances. The green and black paths share identical localization cost, since localization is given an equivalent priority in both instances. The benefits of a lexicographic search are evident in Figure \ref{fig:uav-cost}, as it is clear that  tangible gains in reducing threat exposure, improving localizability, and remaining in communication range can be made by including these quantities in the hierarchy, which leaves higher-priority costs in tact while inducing modest increases in lower-priority costs.

\subsection{Experimental Results with Robot Hardware}

\begin{figure}[ht]
\centering
\includegraphics[width=.9\columnwidth]{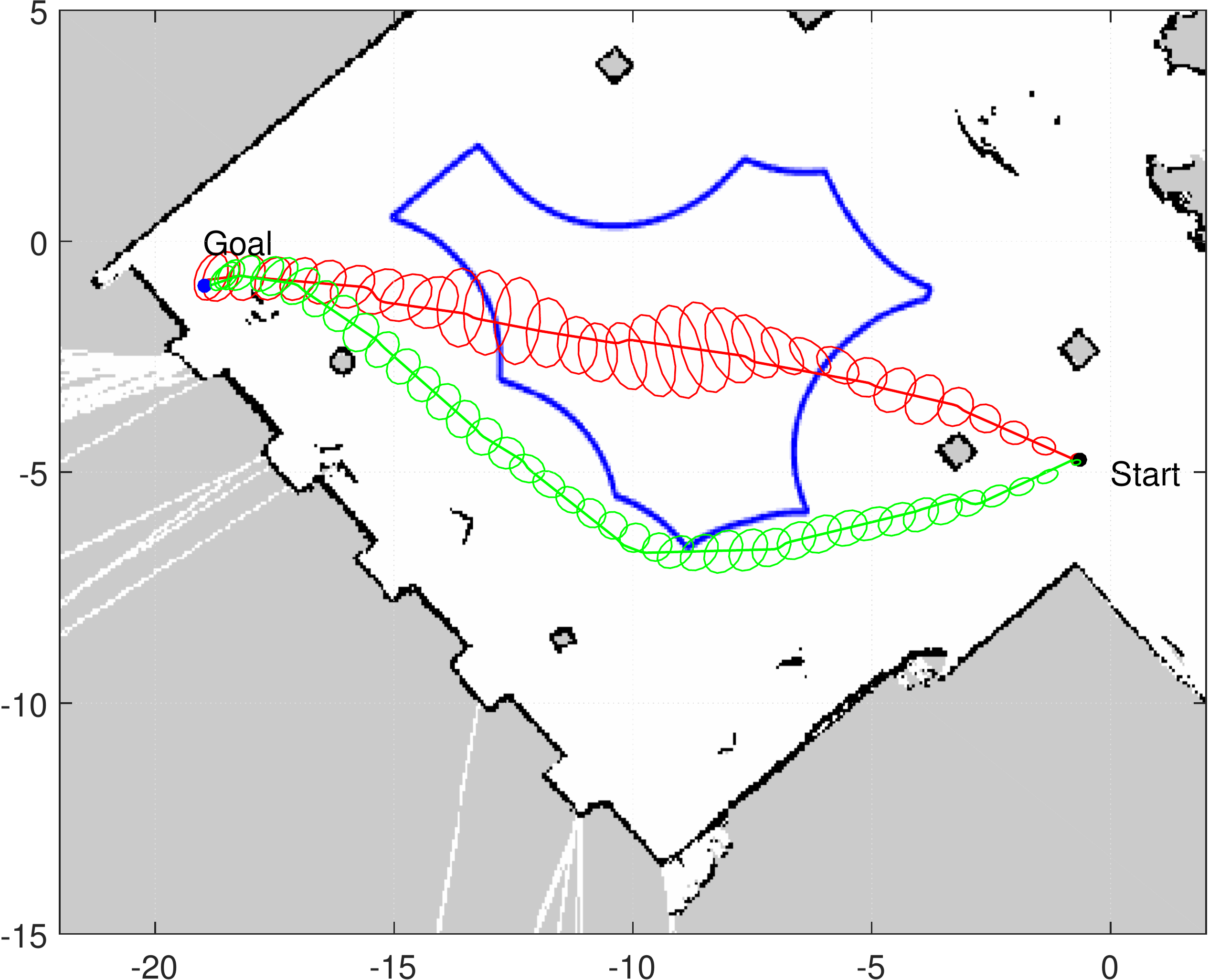}
\caption{Roadmap-derived trajectories executed using a ground robot. The blue line indicates a 3-meter visibility boundary, within which obstacles cannot be observed. The red path and green path are returned by LS when a distance criterion, and localization-distance criteria, are used for planning.}
\label{fig:jackal}
\vspace{-0mm}
\end{figure}

Finally, we implement the proposed algorithm on a mobile robot, the Clearpath Jackal, with motion planning using PRM* subject to Dubins constraints. Our aim is to examine the effectiveness of the lexicographic search method for a real-world mobile robot, in which two criteria, localization cost (primary) and distance cost (secondary), are considered. A Hokuyo UTM-30LX laser scanner, which has a 30 meter range and 270\degree field of view, is mounted on the top of the robot. In order to visualize the benefits of lexicographic search during the mission, Adaptive Monte Carlo Localization (AMCL) \cite{fox}, which uses a particle filter to track the pose of a robot, is employed. We discard laser range returns that are more than 3 meters away from robot, allowing a small-scale indoor environment to produce varied localization outcomes. When the robot is more than 3 meters away from features in the environment, AMCL is forced to rely on wheel odometry only. This "dead-reckoning region" is outlined by a blue line that is shown in Figure \ref{fig:jackal}. The uncertainty of the robot, derived from the output of AMCL, is represented as 95\% confidence ellipses along the plotted trajectories.

Representative robot execution traces of trajectories from the comparison are illustrated in Figure \ref{fig:jackal}. The red trajectory is returned when only the robot's distance traveled is considered. Note that the uncertainty grows dramatically when only odometry information, which is noisy and inaccurate, is available for AMCL. Compared with the shortest distance solution, the lexicographic search solution (green trajectory) stays close to the features so that the growth of localization cost is curbed, without wandering too far from the shortest route. As a result, the lexicographic search solution maintains the uncertainty of robot at a relatively low level. We also note that sometimes, in the course of performing trials, AMCL would fail to localize the robot in the region within the blue line when only a distance criterion is used. In this real-world test, 500 nodes were included in each roadmap constructed. The time required to generate the roadmap and search for a path was found to be less than one second on average per iteration. 

\section{Conclusion}
We have proposed a lexicographic search method intended for use with roadmaps in multi-objective robot motion planning problems, in which competing resources are penalized hierarchically. Over such problems, we have demonstrated that the proposed search method is capable of producing high-quality solutions in efficient runtime to alternative approaches, including a method that employs a single, weighted sum of all competing costs in its objective, and a method that builds and searches an expanded graph whose layers represent the consumption of additional resources. The variant of Dijkstra's algorithm proposed for performing the search offers appealing scalability, as its worst-case complexity scales linearly in the number $K$ of cost criteria. A key benefit of the approach is that, in contrast to planning methods that employ weight coefficients or constraints, no tuning is required, beyond the ordering of cost functions in the hierarchy. Since no constraints other than obstacle avoidance need be imposed, feasible solutions are obtained quickly. Real-world implementation of our method is also verified on a Jackal robot. Future work entails the extension of this method to time-varying costs that are history-dependent, for use in motion planning under uncertainty.




\end{document}